%% file: main.tex
\documentclass[wcp]{jmlr}

\usepackage{longtable}

\usepackage{booktabs}

\usepackage{amsmath,amssymb,amsfonts,mathtools}
\usepackage{algorithm}
\usepackage{algorithmic}
\usepackage{graphicx}
\usepackage{lscape}
\usepackage{subcaption}

\input{mathdef.tex}

\newcommand{\bft}{\fontseries{b}\selectfont}

\usepackage{pifont}
\newcommand{\cmark}{\ding{51}}%

\usepackage{color}

\jmlrvolume{101}
\jmlryear{2019}
\jmlrworkshop{ACML 2019}

\title[Learning to Sample Hard Instances for Graph Algorithms]{Learning to Sample Hard Instances for Graph Algorithms}

 \author{\Name{Ryoma Sato} \Email{r.sato@ml.ist.i.kyoto-u.ac.jp}\\
  \Name{Makoto Yamada} \Email{myamada@i.kyoto-u.ac.jp}\\
  \Name{Hisashi Kashima} \Email{kashima@i.kyoto-u.ac.jp}\\
 \addr 
 Kyoto University, Kyoto 606-8501, Japan. \\
 RIKEN Center for Advanced Intelligence Project, Tokyo 103-0027, Japan.}

\editors{Wee Sun Lee and Taiji Suzuki}

\begin{document}

\maketitle

\begin{abstract}

\textit{Hard instances}, which require a long time for a specific algorithm to solve, help (1) analyze the algorithm for accelerating it and (2) build a good benchmark for evaluating the performance of algorithms. 
There exist several efforts for automatic generation of hard instances. For example, evolutionary algorithms have been utilized to generate hard instances. However, they generate only finite number of hard instances. The merit of such methods is limited because it is difficult to extract meaningful patterns from small number of instances. We seek for a probabilistic generator of hard instances. Once the generative distribution of hard instances is obtained, we can sample a variety of hard instances to build a benchmark, and we can extract meaningful patterns of hard instances from sampled instances.
The existing methods for modeling the hard instance distribution rely on parameters or rules that are found by domain experts; however, they are specific to the problem. Hence, it is challenging to model the distribution for general cases. 
In this paper, we focus on graph problems. We propose \textsc{HiSampler}, the hard instance sampler, to model the hard instance distribution of graph algorithms. \textsc{HiSampler} makes it possible to obtain the distribution of hard instances without hand-engineered features. To the best of our knowledge, this is the first method to learn the distribution of hard instances using machine learning.
Through experiments, we demonstrate that our proposed method can generate instances that are a few to several orders of magnitude harder than the random-based approach in many settings. In particular, our method outperforms rule-based algorithms in the 3-coloring problem.
\end{abstract}

\section{Introduction}
\label{sec: introduction}

Given an algorithm for a combinatorial problem, how do we find instances that take a long time to be solved?
We call such instances \textit{hard instances} \footnote{Hard instances are also referred to as instances that require a long time for \emph{any algorithm} to solve it. In this paper, they are only referred to as instances that require a long time for \emph{a specific algorithm}.}. Finding hard instances is important in algorithm design for the following reasons:

\begin{itemize}
    \item \textbf{Reason 1:} Hard instances help analyze and accelerate the algorithm.
    \item \textbf{Reason 2:} Hard instances help evaluate the performance of other algorithms.
\end{itemize}

\noindent The following illustrative example shows how hard instances help analyze an algorithm: Consider a sorting problem and quicksort, which uses the first element as a pivot. Suppose we do not know that the worst time complexity of quicksort is $\Theta(n^2)$. If we try running quicksort on some random sequences, quicksort seems to solve all instances in $O(n \log n)$ time. However, once a hard instance $[n, n-1, \dots, 3, 2, 1]$ is found, we can see that the worst time complexity is $\Theta(n^2)$. Moreover, such an observation shows that choosing an elaborate pivot improves the algorithm (Reason 1). Besides, such an extreme case is useful for benchmarks because it reveals whether algorithms are robust to worst cases or are efficient only for average cases (Reason 2). Furthermore, in general, if we have hard instances for state-of-the-art algorithms in a benchmark problem instances, we can quickly check whether a new algorithm conquers weakness of the state-of-the-art algorithms (Reason 2).

Finding hard instances is helpful not only for academic subjects but also for practical and industrial subjects. For example, a task scheduler program solves the vertex coloring problem to optimize schedules. However, if a user inputs a malicious schedule (i.e., a hard instance), the scheduler takes a significant amount of time to solve the problem and may hang up. If the developers have such inputs beforehand, they can cope with the issue by setting the appropriate timeout period or maximum size of the input. Another example is the preparation of competitive programming contests. If we create hard instances for each problem, we can accurately check whether the submission is correct or not, which is useful for preparing competitions. Advantages of knowing hard instances are discussed in \cite{SortGA, GA, GATSP} further.

There exist several efforts for automatic generation of hard instances. For example, \cite{SortGA, GA, GATSP} generate hard instances using evolutionary algorithms. However, they generate only finite number of hard instances. The merit of such methods is limited because it is difficult to extract meaningful patterns from small number of instances. We seek for a probabilistic generator of hard instances. Once the generative distribution of hard instances is obtained, we can sample a variety of hard instances from the distribution to build a benchmark, and we can extract meaningful patterns of hard instances from sampled instances.

When we tackle this problem, we must specify the underlying set to model distribution. However, the form of instances depends on the problem. For example, instances are represented by an array in the sorting problem, and they are represented by a set of clauses in the SAT problem. In this paper, we focus on graph problems to fix the underlying set of distributions. Graph problems appear in many important problems. For example, the register allocation problem is formulated as the graph coloring problem \citep{register}, and the maximum clique problem can be utilized for community detection \citep{clique_community}. It motivates us to focus on graph algorithms.

A straightforward method for modeling the hard graph distribution is to use a probabilistic graph model such as the Erd\H{o}s-R\'{e}nyi model \citep{ER}. Though this is generic and simple, this is not efficient because worst time complexity is often far worse than average time complexity \citep{backtrack}. It indicates that, to model the hard graph distribution efficiently, we must develop a method that can capture the structure of the problem and generate \emph{rare} instances. 

In this paper, we propose \textsc{HiSampler}, the \underline{h}ard \underline{i}nstance sampler, to obtain a generative distribution of hard instances of graph algorithms. It models the hard graph distribution using a neural network and trains the model via reinforcement learning. 

Through experiments on seven algorithms of four typical graph problems, we demonstrate that \textsc{HiSampler} can generate instances that are a few to several orders of magnitude harder than the random-based approach in many settings, and that our method outperforms rule-based algorithms in the 3-coloring problem. The implementation of \textsc{HiSampler} is publicly available in \url{https://github.com/joisino/HiSampler} as an open source project.

The major contributions of this paper are as follows:

\begin{itemize}
    \item \textbf{Novel formulation:} We formulate the problem of modeling the hard instance distribution, which is practically important.
    \item \textbf{Novel method:} We propose \textsc{HiSampler}, an effective method to model the hard instance distribution for a given graph algorithm.
    \item \textbf{Experimental evidence:} We demonstrate the effectiveness of \textsc{HiSampler} through extensive experiments using seven algorithms of four problems.
\end{itemize}

Table \ref{tab: summary} contrasts \textsc{HiSampler} against other methods for generating hard instances.

\begin{table*}[tb]
    \caption{Qualitative comparison with other methods for generating hard instances.}
    \vspace{-0.1in}
    \centering
    \begin{tabular}{lcccc} \toprule
    & Random & Rule-based & Generic Algorithm & \textsc{HiSampler} \\ \midrule
    Effective &  & \cmark & \cmark & \cmark \\
    Without hand-engineering & \cmark & & \cmark & \cmark \\
    Problem-agnostic & \cmark & & \cmark & \cmark \\
    Distribution & \cmark & & & \cmark \\
    Sample-efficient & & \cmark & & \cmark \\ 
    \bottomrule
    \end{tabular}
    \label{tab: summary}
    \vspace{-.1in}
\end{table*}

\section{Proposed Method}
We first describe the problem setting of this paper. Then, we propose \textsc{HiSampler}, an effective method to learn the distribution of hard instances for graph algorithms.
\subsection{Problem Setting} \label{sec: setting}

We specify the task of learning the hard instance distribution of graph algorithms.
In particular, we develop a method that models the hard instance distribution for algorithms for undirected, unweighted, and simple graphs. It is because they include many important problems. For example, the register allocation problem is formulated as the graph coloring problem \citep{register} and the maximum clique problem can be utilized for community detection \citep{clique_community}.

We aim to develop a method that relies on no problem specific properties; instead, we use only the hardness measures of the problem instances: $\textrm{hardness}(\boldA, L)$, where $L$ is the given algorithm, $\boldA \in \{0,1\}^{n(n-1)/2}$ is the adjacency matrix, and $n$ is the number of vertices.
The design of the hardness value is arbitrary if it can be obtained by actually running the algorithm on the instance. For example, in our experiment, the hardness is measured by the number of recursive calls DSATUR \citep{brelaz} makes to solve the instance and by the real time Nauty \citep{Nauty} spends to solve the instance.

Formally, given an algorithm $L$, we aim to develop a method that models a generative distribution $\mathcal{D}_L$ of graphs that maximizes $\mathbb{E}_{\boldA \sim \mathcal{D}_L}[\text{hardness}(\boldA, L)]$. Besides, we develop a method that satisfies the following key assumptions.

\vspace{.05in}
\noindent {\bf Assumption1. Small instance:} We fix the number of vertices $n$ because we can generate arbitrarily hard instances just by increasing the number of vertices, which is not practical. Moreover, since small instances can be visualized and are easy to interpret and analyze, it is important to generate hard instances without increasing the size of the instance. 

\vspace{.05in}
\noindent {\bf Assumption2. Sample efficiency:} Evaluating the hardness value is time consuming, especially when the instance is hard. Besides, algorithms that require special devices such as GPU and multiple cores cost much even if they run in a short period of time.
Therefore, we cannot evaluate too many instances, which motivates us to find hard instances more efficiently. To overcome this problem, we set the budget $B$ of evaluation. In other words, we do not evaluate more than $B$ instances during training. It should be noted that evolutionary algorithms are not sample-efficient because they evaluate the fitness functions of large population in each iteration.

Table \ref{tab: notations} summarizes the notations we use throughout the paper.

\begin{table*}[tb]
    \caption{Notations.}
    \vspace{-0.1in}
    \centering
    \begin{tabular}{cl} \toprule
    Notation & Description \\ \midrule
    $G$ & A graph (i.e., an instance) \\
    $V$ & The whole set of nodes \\
    $n$ & The number of nodes (i.e., $n = |V|$)\\
    $\boldA \in \{0, 1\}^{n(n-1)/2} $ & The upper triangular part of an adjacency matrix \\
    $\boldP \in [0, 1]^{n(n-1)/2}$ & The upper triangular part of the a probabilistic adjacency matrix \\
    $L$ & A graph algorithm \\
    $\text{hardness}(\boldA, L) \in \mathbb{R}$ & The hardness value of graph $A$ for algorithm $L$ \\ 
    $B$ & The maximum number of evaluations \\
    $l$ & The number of layers of the neural network \\
    $d_i ~(i = 0, 1, \dots, l)$ & The dimensions of the hidden layers of the neural network \\ 
    \bottomrule
    \end{tabular}
    \label{tab: notations}
    \vspace{-.1in}
\end{table*}

\subsection{Hard Instance Sampler}
\begin{figure}[tb]
    \centering
    \includegraphics[width=\hsize]{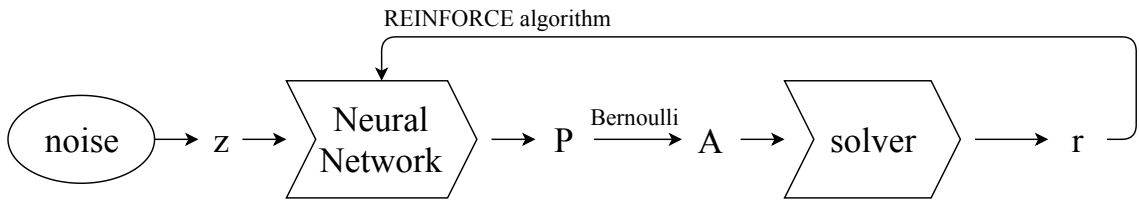}
    \caption{The overview of \textsc{HiSampler}}
    \label{fig: overview}
\end{figure}

We propose \textsc{HiSampler} to model the distribution of hard instances of graph algorithms. Figure \ref{fig: overview} illustrates the overview of \textsc{HiSampler}.

\vspace{.05in}
\noindent \textbf{Probabilistic Model:} We consider distributions on the binary vector $\{0, 1\}^{n(n-1)/2}$ because a graph $G$ is represented by an adjacency matrix $\boldA \in \{0, 1\}^{n(n-1)/2}$. \textsc{HiSampler} models the distribution by a fully-connected neural network $N$ with parameters $\boldtheta$. Let $l$ denote the number of layers of $N$ and let $d_i$ be the dimensions of the hidden layers of $N$. $d_0$ is the input dimension and $d_l = n(n-1)/2$ is the output dimension. The neural network $N$ takes a noise $\boldz \sim \mathcal{N}(0, I_{d_0})$ from the standard normal distribution as input and outputs a probabilistic adjacency matrix $\boldP \in [0, 1]^{n(n-1)/2}$. Then, an adjacency matrix $\boldA$ is sampled from $\text{Bernoulli}(\boldP)$. Namely, $Pr[\boldA_i \mid \boldP] = \boldP_i ~(i = 1, 2, \dots n(n-1)/2)$ and each dimension of $\boldA$ is conditional independent given $\boldP$. It is worth noting that $\boldA_i$ and $\boldA_j ~(i \neq j)$ are not independent without any condition because $\boldP_i$ and $\boldP_j$ are not independent. Therefore, \textsc{HiSampler} can model nonlinear relationships between edges.

\vspace{0.05in}
\noindent \textbf{Optimization:} \textsc{HiSampler} optimizes the parameters of the neural network $N$ using immediate-reinforcement learning. In this framework, the environment gives a noise $\boldz$ to the agent, the agent generates a graph $\boldA$ as an action, and the cost of solving the instance is fed back to the agent as a reward $r$ (i.e., $r = \textrm{hardness}(\boldA, L)$). The policy of the agent is modeled by the neural network $N$. We optimize the parameters of the neural network $N$ using REINFORCE algorithm~\citep{williams}:
\[ \boldtheta \leftarrow \boldtheta + \alpha r \frac{\partial}{\partial \boldtheta} \sum_{i=1}^{n(n-1)/2} (\log \boldP_i^{\boldA_i} + \log (1 - \boldP_i)^{(1 - \boldA_i)}), \]
where $\alpha$ is the learning rate. The procedure used to train the neural network model is shown in Algorithm \ref{algo: vanilla}.

\vspace{0.05in}
\noindent \textbf{Prioritized Experience Replay:} The main challenge of learning the distribution of hard instances is that hard instances are sparse in the instance space. This tendency is significant especially for sophisticated algorithms. To alleviate this issue, we use a variant of prioritized experience replay \citep{PER}. Namely, we maintain an experience pool that contains top-$K$ hard instances. In each iteration, we sample an instance from the pool, and we train the model using the sampled experience. We refer to \textsc{HiSampler} with prioritized experience replay as \textsc{HiSampler-PER} to distinguish it from \textsc{HiSampler-vanilla}. The training procedure of \textsc{HiSampler-PER} is shown in Algorithm \ref{algo: PER}. We simplify the original prioritized experience replay by (1) using uniform distribution of top-$K$ instances instead of weighted sampling and (2) using reward to prioritize experiences instead of TD-Error. Though this is simple, we found this works well in practice for \textsc{HiSampler}. We empirically demonstrate effectiveness of this method in Section \ref{sec: experiments}.

\begin{algorithm}[tb]
\caption{\textsc{HiSampler-vanilla}}
\label{algo: vanilla}
\begin{algorithmic}[1]
\REQUIRE Algorithm $L$; Budget $B$.
\ENSURE Hard graph distribution.
\STATE Initialize the parameters of the neural network $N$.
\STATE $r_{best} \leftarrow 0$ 
\FOR{$B$ iterations}
\STATE $\boldz \leftarrow$ a sample from $\mathcal{N}(0, I_d)$
\STATE $\boldP \leftarrow N(\boldz; \boldtheta)$
\STATE $\boldA \sim \textrm{Bernoulli}(\boldP)$
\STATE $r \leftarrow \textrm{hardness}(\boldA, L)$
\STATE $\boldtheta \leftarrow \boldtheta + \alpha r \frac{\partial}{\partial \boldtheta} \sum_{i=1}^{n(n-1)/2} (\log \boldP_i^{\boldA_i} + \log (1 - \boldP_i)^{(1 - \boldA_i)})$ \hfill $\rhd$ REINFORCE algorithm
\ENDFOR
\STATE \textbf{return} $N$
\end{algorithmic}
\end{algorithm}

\begin{algorithm}[tb]
\caption{\textsc{HiSampler-PER}}
\label{algo: PER}
\begin{algorithmic}[1]
\REQUIRE Algorithm $L$; Budget $B$; Pool size $K$.
\ENSURE Hard graph distribution.
\STATE Initialize the parameters of the neural network $N$.
\STATE $r_{best} \leftarrow 0$ 
\STATE $Q \leftarrow$ an empty priority queue
\FOR{$B$ iterations}
\STATE $z \leftarrow$ a sample from $\mathcal{N}(0, I_d)$
\STATE $\boldP \leftarrow N(\boldz; \boldtheta)$
\STATE $\boldA \sim \textrm{Bernoulli}(\boldP)$
\STATE $r \leftarrow \textrm{hardness}(\boldA, L)$
\STATE $Q\text{.push(}(r, \boldz, \boldA)\text{)}$ \hfill $\rhd$ Add a new sample to the queue
\IF{size($Q$) $> K$}
    \STATE $Q\text{.pop()}$ \hfill $\rhd$ Remove the least significant sample
\ENDIF
\STATE $(r', \boldz', \boldA') \leftarrow$ a uniform random sample from $Q$.
\STATE $\boldP' \leftarrow N(\boldz'; \boldtheta)$
\STATE $\boldtheta \leftarrow \boldtheta + \alpha r \frac{\partial}{\partial \boldtheta} \sum_{i=1}^{n(n-1)/2} (\log \boldP'_i{}^{\boldA'_i} + \log (1 - \boldP'_i)^{(1 - \boldA'_i)})$
\ENDFOR
\STATE \textbf{return} $N$
\end{algorithmic}
\end{algorithm}

\subsection{Complexity Analysis} \label{sec: complexity}

We analyze the time complexity of \textsc{HiSampler}. The bottleneck step of \textsc{HiSampler} is forward and backward calculation of the neural network. It takes $O(\sum_{i = 0}^{l-1} d_i d_{i+1})$ time per iteration. Therefore, the time complexity of \textsc{HiSampler} is $O(n^2)$ with respect to the graph size because the dimension of the output layer is $d_l = n(n-1)/2$. The additional computation needed for \textsc{HiSampler-PER} is maintaining the priority queue. It takes $O(\log K)$ time per iteration, which is negligibly small in practice.

\section{Extensions}

In our proposed method, the choice of the hardness function is arbitrary. Therefore, \textsc{HiSampler} can find not only instances that take a long time to be solved but also hard instances in terms of other criteria. We introduce two important examples.

\subsection{Estimating Approximation Ratio}
Let $L$ be an approximation algorithm, let $\boldA$ be an instance of the problem, let $L(\boldA)$ be the object value of the solution that $L$ outputs for $\boldA$, and let $\text{OPT}(\boldA)$ be the optimal objective value of an optimal solution of $\boldA$. The approximation ratio of $L$ is defined by $\displaystyle r(L) = \max_{\text{$\boldA$ is an instance}} \frac{L(\boldA)}{\text{OPT}(\boldA)}$ for a minimizing problem and by $\displaystyle r(L) = \max_{\text{$\boldA$ is an instance}} \frac{\text{OPT}(\boldA)}{L(\boldA)}$ for a maximizing problem. Estimating the approximation ratios is important for investigating the performance of the approximation algorithms. However, it is not trivial what instance maximizes the term. Here, we use $\frac{L(\boldA)}{\text{OPT}(\boldA)}$ or $\frac{\text{OPT}(\boldA)}{L(\boldA)}$ as the hardness value of the instance $\boldA$; then, we can search the maximizer by \textsc{HiSampler}. We will show an illustrative example with the well-known minimum vertex cover algorithm in Section \ref{sec: vcapp}.

\subsection{Hard Instances for Enumerating Algorithms}

Enumerating algorithms output all the elements that satisfy some property. The amortized time and maximum delay are sometimes investigated for evaluating the efficiency of enumerating algorithms. \textsc{HiSampler} can generate hard instances in terms of the amortized time and maximum delay by setting these measures as the hardness value.

\section{Related Work}

\vspace{.05in}
\noindent {\bf Constructing Hard Instances:} There have been several researches on constructing hard instances of combinatorial problems.
Hard instance generation was first studied in relation to the phase transition phenomena \citep{cheeseman, hogg}, which utilizes order parameters to generate hard instances.
The three coloring instance generation by Mizuno and Nishihara~(\citeyear{mizuno}) and the graph isomorphism instance generation by Neuen et al.~(\citeyear{neuen}) generate hard instances with rule-based algorithms. However, these works depend on problem specific knowledge, whereas our method is independent of the problem. 
Another approach is to generate instances by reducing other related problems. For example, the Latin square problem is found useful for constructing a benchmark for the  graph-coloring problem \citep{Gomes} and SAT \citep{achlioptas}. However, the conversion of an instance of the Latin square problem into those of other problems also requires problem-specific knowledge. 
The methods that are most related to this paper are evolutionary algorithms \citep{SortGA, GA, GATSP}. They optimize the hardness of instances using the evolutionary algorithm. However, they require designing gene representation for each task and cannot find a distribution but find a finite set of hard instances.

\vspace{.05in}
\noindent {\bf Deep Generative Graph Models:}
Recently, several generative graph models utilizing deep learning techniques have been proposed. 
The variational graph auto-encoder \citep{GAE} is one of the first models of this kind. It is a variant of the Variational Auto Encoder (VAE), which outputs a probabilistic adjacency matrix. This model was used for the link prediction of citation networks.
Then, VAE \citep{GraphVAE,Graphite,ma}, Generative Adversarial Networks (GAN) \citep{GraphGAN,NetGAN}, and sequential generation \citep{GraphRNN,CGVAE,GCPN} based generating models were proposed. In particular, they succeeded in generating various de-novo chemical materials and modeling real-world networks. ORGAN \citep{ORGAN} utilizes SeqGAN \citep{SeqGAN} and reinforcement learning to generate molecular graphs with the desired properties. It uses SMILES \citep{SMILES} to represent a molecular graph because SeqGAN generates a sequence of symbols rather than a graph itself. MolGAN \citep{MolGAN} is another graph generative model utilizing GAN and reinforcement learning. It models the probabilistic adjacency matrix and attributes of graphs directly instead of using SMILES.
The differences between \textsc{HiSampler} and deep generative graph models are (1) these models use training data that contain graphs with high objective values whereas \textsc{HiSampler} uses no training data and (2) many of these models are designed for generating molecular graphs, where the size of graphs is typically at most dozens, whereas \textsc{HiSampler} can generate graphs with more than a hundred nodes.

\section{Experiments}
\label{sec: experiments}

We will answer the following questions through experiments:
\begin{description} \setlength{\itemsep}{0.05in} \setlength{\parskip}{0in}
\item[Q1.] \textbf{Scalability:} How fast is \textsc{HiSampler}?
\item[Q2.] \textbf{Effectiveness:} Does \textsc{HiSampler} generate harder instances than existing methods?
\item[Q3.] \textbf{Knowledge Extraction:} Can \textsc{HiSampler} provide insights for algorithm design?
\item[Q4.] \textbf{Diversity:} Is the distribution of \textsc{HiSampler} rich in diversity?
\item[Q5.] \textbf{Extensions:} Can \textsc{HiSampler} estimate an approximation ratio?
\item[Q6.] \textbf{Effective Patterns:} How can we extract effective patterns from the obtained hard graph distribution?
\end{description}

\vspace{0.05in}
\noindent \textbf{Common Experimental Setup: }
We set the number of layers of \textsc{HiSampler} to three and the dimensions of the hidden layers to $d_0 = 10$, $d_1 = 100$, $d_2 = 500$, and $d_3 = n(n-1)/2$ throughout experiments. The activation functions in the hidden layers are ReLU, and the final output is processed by sigmoid activation. We use Adam \citep{adam} with learning rate $0.0001$ to train the model. We set the pool size of \textsc{HiSampler-PER} to $K = 10$ throughout experiments. We conduct experiments with Intel Xeon E5-2690 CPU. It should be noted that we can speed up the computation of \textsc{HiSampler} by GPUs, but we do not use GPUs for fair comparison.

\subsection{Scalability}
\label{sec: scalability}

As we mentioned in Section \ref{sec: complexity}, the complexity of \textsc{HiSampler} is $O(n^2)$.
We investigate time consumption of training and sampling of \textsc{HiSampler} through experiments. We sample $100$ instances from each of $10$ \textsc{HiSampler}s, and we execute one step of training for each sample. We omit the time of evaluating the hardness value during training because the overhead of evaluation is common with other methods. Furthermore, we consider that training is already done when the evaluation time overwhelms model computation. If the evaluation takes much time in the initial evaluation, we should make the graph size smaller because generating small instances is important (Assumption 1 in Section \ref{sec: setting}).

\begin{figure}[tb]
 \begin{minipage}{0.48\hsize}
\begin{center}
\vspace{-0.25in}
\includegraphics[width=\hsize, height=2in, keepaspectratio]{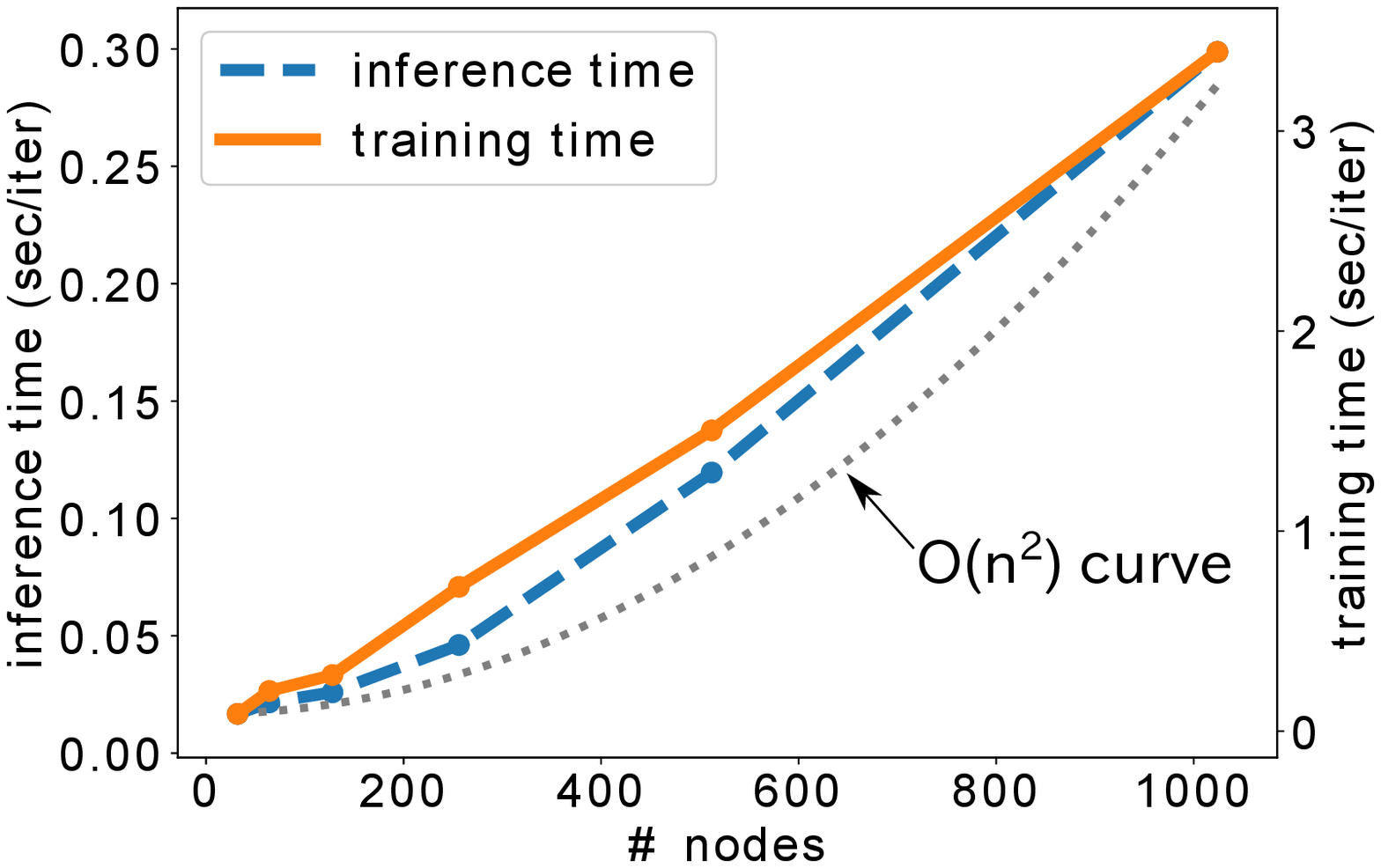}
\end{center}
\vspace{-0.1in}
\caption{The time consumption of the inference and training processes of \textsc{HiSampler}.}
\label{fig: scalability}
 \end{minipage} \hfill
 \begin{minipage}{0.48\hsize}
\begin{center}
\includegraphics[width=\hsize, height=2in, keepaspectratio]{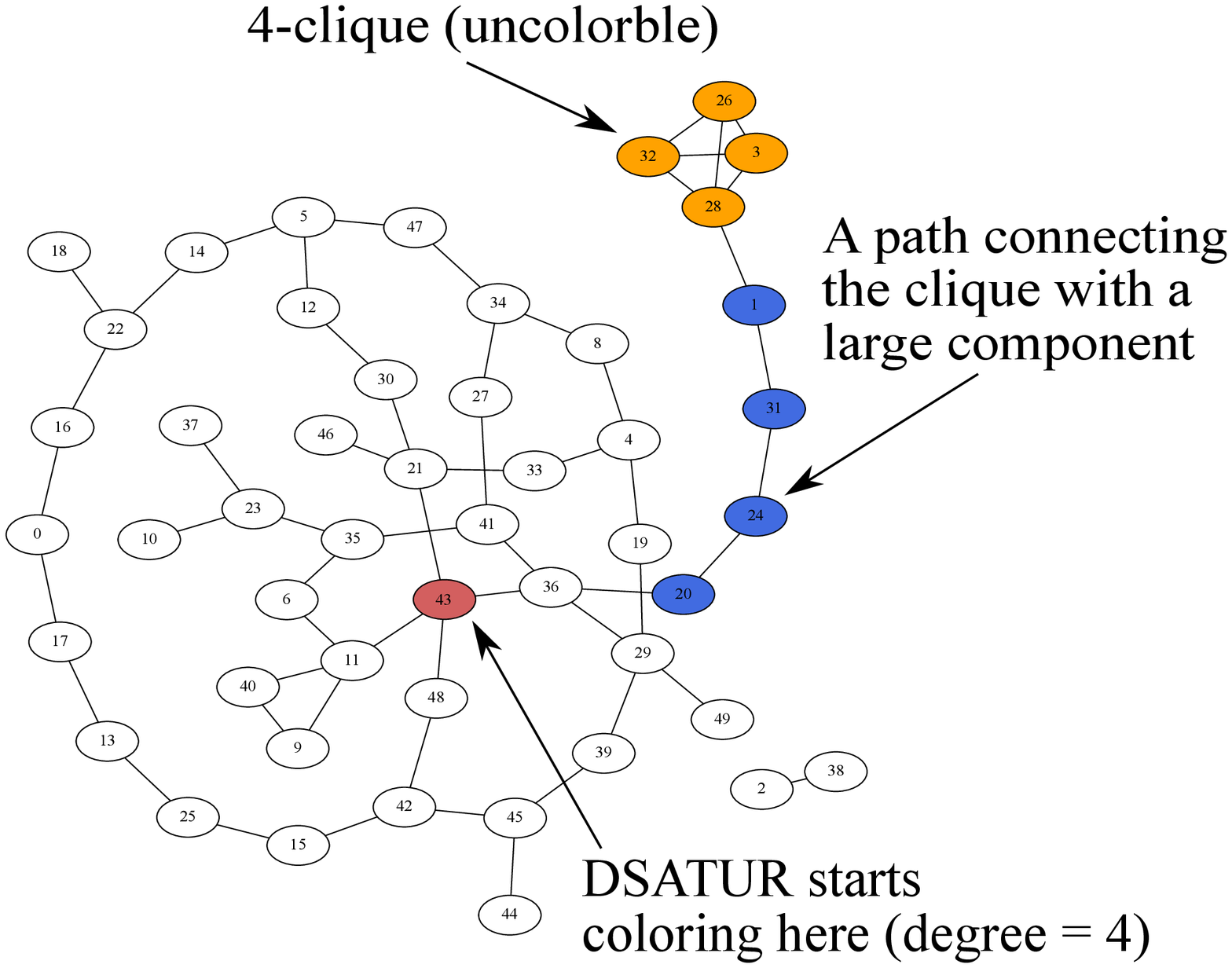}
\end{center}
\vspace{-0.12in}
\caption{An example of 3-coloring instances that \textsc{HiSampler} generates. DSATUR takes more than one billion recursive calls to solve this instance.}
\label{fig: brelaz_example}
 \end{minipage}
 \vspace{-0.2in}
\end{figure}

Figure \ref{fig: scalability} reports the mean time of a single iteration of sampling and training. This shows that the computation does not grow much even if the number of nodes increases. In particular, one iteration of the training takes only four seconds even with $1024$ nodes. It indicates that \textsc{HiSampler} is highly efficient.

\subsection{Effectiveness}

We demonstrate how hard instances \textsc{HiSampler} can generate compared to existing methods. We use the 3-coloring problem, the minimum vertex cover problem, the maximum clique problem, and the graph isomorphism problem, and the following seven algorithms for these problems

\vspace{0.05in}
\noindent \textbf{DSATUR (3-coloring):} This is a backtracking search method based on DSATUR \citep{brelaz}. 
It assigns colors to the vertices one by one. 
At each step, it chooses one of the uncolored vertices that have the least number of candidate colors. If there are many such vertices, it chooses the vertex with the maximum degree.
If the color assignment becomes inconsistent, it backtracks until it finds a consistent assignment. This method always outputs exact solution whereas the original DSATUR is not. We use the number of recursive calls as the hardness value.

\vspace{0.05in}
\noindent \textbf{MiniSat (3-coloring):} This reduces the 3-coloring problem to the SAT problem, and this solves the reduced instance using MiniSat \citep{Minisat}. We use the number of decisions MiniSat reports as the hardness value.

\vspace{0.05in}
\noindent \textbf{B\&B (minimum vertex cover):} This is a branch and bound algorithm that uses a greedy maximal matching as an upper bound. We use the number of recursive calls as the hardness value.

\vspace{0.05in}
\noindent \textbf{BK (maximum clique):} This is a branch and bound algorithm based on the Bron-Kerbosch method \citep{BK}. This prunes the state when the union of the selected nodes and candidate nodes is smaller than the maximum clique found so far. Note that the original Bron-Kerbosch method enumerates all the maximal cliques whereas this algorithm only outputs a maximum clique. We use the number of recursive calls as the hardness value.

\vspace{0.05in}
\noindent \textbf{MCS (maximum clique):} This is MCS \citep{MCS}, a branch and bound algorithm. We use the time consumption as the hardness value ($10^{-2}$ sec).

\vspace{0.05in}
\noindent \textbf{FMC (maximum clique):} This is Fast Max-Clique Finder \citep{fmc}, a hierarchical-pruning based algorithm. We use the time consumption as the hardness value ($10^{-6}$ sec)

\vspace{0.05in}
\noindent \textbf{Nauty (graph isomorphism):} This is Nauty \citep{Nauty}, one of the state-of-the-art graph isomorphism solvers. We use the time consumption as the hardness value ($10^{-6}$ sec) \\

To compare the effectiveness of \textsc{HiSampler}, we use the following baseline methods.

\vspace{0.05in}
\noindent \textbf{Generic algorithm:} This searches hard instances using an evolutionary algorithm. We use the adjacency matrix $A$ as gene representation. We use the same hyperparameters as \cite{GA}. Namely, the population size is $30$, crossover is performed uniformly, mutation occurs with uniform probability with adapting mutation rate, and the fitness is the hardness value.

\vspace{0.05in}
\noindent \textbf{Random graphs:} This samples $B$ graphs from the Erd\H{o}s-R\'{e}nyi model \citep{ER} and reports the hardest one.

\vspace{0.05in}
\noindent \textbf{Rule-based:} We use several rule-based methods in the 3-coloring problem and the graph isomorphism problem. \cite{cheeseman} and \cite{hogg} used the Erd\H{o}s-R\'{e}nyi model with carefully tuned parameters for the 3-coloring problem (i.e., $p = \frac{4.6}{n-1}$ and $p = \frac{3.4}{n-1}$, respectively). \cite{vlasie} found that a regular structure plays a key role for hard instances and generated graphs with less 3-paths for the 3-coloring problem. \cite{mizuno} and \cite{neuen} used characteristic gadgets to construct hard instances. We generate $B$ graphs using these methods and report the hardest one.

The most important step for \textsc{HiSampler} and the generic algorithm is initialization. It is known that the algorithm takes long time for graphs with certain range of edge density and that it takes short time for graphs with other density \citep{cheeseman}. For example, 3-coloring algorithms can easily assert there is no solutions for dense graphs. If the initial distribution of \textsc{HiSampler} or initial population of the generic algorithm is far from the hard region, it takes much time for them to generate hard instances. To alleviate this problem, we determine the edge probability $p^*$ beforehand where each algorithm takes long time to process graphs with this density. We can use prior knowledge about the algorithm to determine $p^*$. Alternatively, if we do not have such knowledge, we evaluate random instances of different edge probabilities (e.g., $p = 0.01, 0.02, \dots, 0.99$), and we can use the hardest one as $p^*$. It consumes negligibly small budget. We initialize the population of the generic algorithm by Erd\H{o}s-R\'{e}nyi model with $p = p^*$, and we initialize the bias of the last layer of \textsc{HiSampler} as $b = - \log(\frac{1}{p^*}  - 1)$ so that $\sigma(b) = p^*$, where $\sigma$ is the sigmoid function. We initialize the weight matrices of \textsc{HiSampler} with Xavier initializer \citep{xavier} and biases of the lower layers with zeros as the default setting of the library. It should be noted that the other hyperparameters than the graph size $n$ and the initial edge probability $p^*$ are fixed throughout experiments.

We set the budget size as $B = 10000$, and we stop a method when it takes more than a day. We measure the hardness value of the hardest instance each method finds. We run 5 experiments for each method with different seeds and we report the mean of 5 runs. Table \ref{table: results} summarizes the result of the experiments. We can see the following observations.

\vspace{0.05in}
\noindent \textbf{Observation 1. Prioritized experience replay is effective:} \textsc{HiSampler-PER} consistently outperforms \textsc{HiSampler-vanilla} except for FMC, where \textsc{HiSampler-vanilla} slightly outperforms \textsc{HiSampler-PER}. It indicates that prioritized experience replay works well for \textsc{HiSampler}.

\vspace{0.05in}
\noindent \textbf{Observation 2.} \textsc{HiSampler-PER} \textbf{outperforms the generic algorithm:} \textsc{HiSampler-PER} consistently outperforms the generic algorithm especially in DSATUR algorithm. It shows that \textsc{HiSampler} learns the hard distribution effectively. 

\vspace{0.05in}
\noindent \textbf{Observation 3.} \textsc{HiSampler} \textbf{outperforms random-based methods:} \textsc{HiSampler} consistently outperforms the Erd\H{o}s-R\'{e}nyi model with $p = p^*$. It demonstrates that the distribution of \textsc{HiSampler} is not random. \textsc{HiSampler} learns effective structure of the hard graph distribution.

\vspace{0.05in}
\noindent \textbf{Observation 4.} \textsc{HiSampler} \textbf{outperforms rule-based methods:} \textsc{HiSampler} consistently outperforms rule-based methods. It indicates that \textsc{HiSampler} can find highly effective structure for hard instances that could not be found manually.

It should be noted that we measured the hardest instance that each method found because the generic algorithm aims at searching a hard instance instead of modeling the hard instance distribution. We will investigate the properties of the distribution (e.g., diversity, mining patterns) in the later experiments.

\begin{table*}[tb]
\caption{The experimental results}
\vspace{-0.2in}
\begin{center}
\hspace{0.2in} \begin{tabular}{lrrrr}
\hline \hline
Problem & \multicolumn{2}{c}{3-coloring} & \multicolumn{1}{c}{Vertex Cover} \\
\cmidrule(lr{1.0em}){2-3}
Algorithm & \multicolumn{1}{c}{DSATUR} & \multicolumn{1}{c}{MiniSat} & \multicolumn{1}{c}{B\&B} \\
n                                        & 50                & 200          & 50 \\
$p^*$                                    & 0.1               & 0.025        & 0.1\\ \hline
\textsc{HiSampler-vanilla}               & 261331027.8       & 1120.4       & 8145.2  \\ 
\textsc{HiSampler-PER}                   & \bft{610024238.8} & \bft{2674.2} & \bft{21376.4} \\ \hline
Generic Algorithm                        & 2464.8            & 660.8        & 8127.6        \\
Erd\H{o}s-R\'{e}nyi $p = p^*$            & 407.0             & 693.6        & 3259.2        \\
Erd\H{o}s-R\'{e}nyi $p = 0.1$            & 407.0             & 351.8        & 3259.2        \\
Erd\H{o}s-R\'{e}nyi $p = 0.5$            & 2.0               & 282.4        & 2227.8        \\
Erd\H{o}s-R\'{e}nyi $p = 0.9$            & 2.0               & 276.0        & 1160.8        \\
Cheeseman et al. (\citeyear{cheeseman})  & 597.8             & 810.2        & N/A           \\
Hogg and Williams (\citeyear{hogg})      & 3883.8            & 815.8        & N/A           \\
Vlasie (\citeyear{vlasie})               & 240867.8          & 708.2        & N/A           \\
Mizuno and Nishihara (\citeyear{mizuno}) & 166294.6          & 875.4        & N/A           \\
\hline
\end{tabular}
\newline
\vspace*{0.05in}
\newline
\begin{tabular}{lrrrr}
\hline \hline
Problem & \multicolumn{3}{c}{Clique} & \multicolumn{1}{c}{Isomorphism} \\
\cmidrule(lr{1.0em}){2-4}
Algorithm & \multicolumn{1}{c}{BK} & \multicolumn{1}{c}{MCS} & \multicolumn{1}{c}{FMC} & \multicolumn{1}{c}{Nauty} \\
n                                        & 32             & 150              & 32              & 50 \\
$p^*$                                    & 0.9            & 0.9              & 0.9             & 0.9 \\ \hline
\textsc{HiSampler-vanilla}               & 65460.4        & 573.4        & \bft{6119052.0} & 604.0 \\ 
\textsc{HiSampler-PER}                   & \bft{110591.0} & \bft{1877.0} & 5682186.0       & \bft{786.0} \\ \hline
Generic Algorithm                        & 25019.6        & 1117.4       & 2588840.0       & 57.6 \\
Erd\H{o}s-R\'{e}nyi $p = p^*$            & 6380.8         & 276.2        & 58655.6         & 10.2 \\
Erd\H{o}s-R\'{e}nyi $p = 0.1$            & 82.8           & 0.0          & 3259.2          & 10.5  \\
Erd\H{o}s-R\'{e}nyi $p = 0.5$            & 502.8          & 1.0          & 5284.2          & 2.7  \\
Erd\H{o}s-R\'{e}nyi $p = 0.9$            & 6380.8         & 276.2        & 58655.6         & 10.2 \\
$R(B(G_n, \sigma))$ (\citeyear{neuen})   & N/A            & N/A          & N/A             & 230.0 \\
shrunken multipedes (\citeyear{neuen})   & N/A            & N/A          & N/A             & 102.6 \\
\hline
\end{tabular}
\end{center}
\label{table: results}
\vspace{-.2in}
\end{table*}

\subsection{Knowledge Extraction} \label{sec: extraction}

We demonstrate how a hard instance provides helpful insight into making a better search algorithm using a concrete example. Figure \ref{fig: brelaz_example} shows an example of the instances \textsc{HiSampler} generates. 
There are no solutions for this instance because it has a 4-clique $C$ (highlighted in orange).
However, DSATUR cannot explicitly detect 4-cliques. 
It first assigns colors to $V \backslash C$. 
Every time it finds a solution for $V \backslash C$, the partial solution is immediately rejected when the algorithm starts to color the 4-clique $C$. 
Then, the search is back-tracked and the algorithm starts to find other assignments of $V \backslash C$. 
However, it does not obtain any result because any assignment will be rejected by the 4-clique $C$.
Finally, the algorithm finds all the valid assignments of $V \backslash C$, and reports that there are no solutions for this instance.
The key point is that the 4-clique $C$ is connected to $V \backslash C$ by a path $P$ (highlighted in blue). $P$ plays a role of a ``bottleneck''. When the algorithm is coloring $V \backslash (C \cup P)$, the number of color candidates of vertices of $P$ are at most two, and the degrees of them are only two. Therefore, DSATUR is reluctant to color these vertices.
From this analysis, we can improve the backtracking search by preprocessing: deleting vertices whose degree is not more than two. Deleting such vertices does not change the answer because we can color the vertices whose degree is not more than two whatever the coloring assignment of the other vertices is: just color the vertex with the color that is not the same as the colors of adjacent nodes. This improvement helps avoid the problem described above.
This discussion is a good example to show that analyzing a hard instance helps design an algorithm robust to hard instances (corresponds to Reason 1 in Section \ref{sec: introduction}).

\subsection{Diversity}
\label{sec: diversity}

We show that a variety of hard instances are sampled from the distribution \textsc{HiSampler} learns. Diversity of hard instances helps build a benchmark and extract meaningful pattern. We train \textsc{HiSampler-vanilla} for DSATUR algorithm. The hardness value of the hardest instance $x^*$ that \textsc{HiSampler} finds is $1637666819$. We sample $1000$ instances from the distribution for which the Jaccard indices of edges between $x^*$ are less than $0.7$. The mean of the hardness values of these instances is $3943028.974$, which is still harder than the random models and the rule-based methods, and the mean of the Jaccard indices is $0.646$. Moreover, the hardness value of the hardest instance among them is $819309215$, keeping the Jaccard index $0.694$. It shows that \textsc{HiSampler} retains diversity, whereas the genetic algorithm only generates a fixed number of instances.

\subsection{Extensions}
\label{sec: vcapp}

We show an illustrative example to estimate the approximation ratio using the greedy algorithm for the minimum vertex cover problem. It is known that the approximation ratio of the greedy algorithm is $2$. We use the Erd\H{o}s-R\'{e}nyi model with $p = 0.1$ and \textsc{HiSampler-vanilla} with $p^* = 0.1$. We set the number of vertices as $n = 50$. The other settings are common with previous experiments. We use $r(\boldA) = \exp(10 \cdot L(\boldA) / \text{OPT}(\boldA))$ as the hardness value, which is monotonically increasing for $L(\boldA) / \text{OPT}(\boldA)$. We found that the slope of $L(\boldA) / \text{OPT}(\boldA)$ is too gentle to train the model, and used the objective function instead. We ran $5$ experiments with difference seeds. None of the uniformly random graph models found an instance $\boldA$ that satisfies $L(\boldA) / \text{OPT}(\boldA) = 2$. However, all five \textsc{HiSampler} models succeeded in finding an instance $\boldA$ that satisfies $L(\boldA) / \text{OPT}(\boldA) = 2$. It shows that \textsc{HiSampler} is useful for estimating the approximation ratios of approximation algorithms.

\subsection{Effective Patterns}

\begin{figure}[tb]
\begin{minipage}{0.19\hsize}
\begin{center}
\includegraphics[width=\hsize]{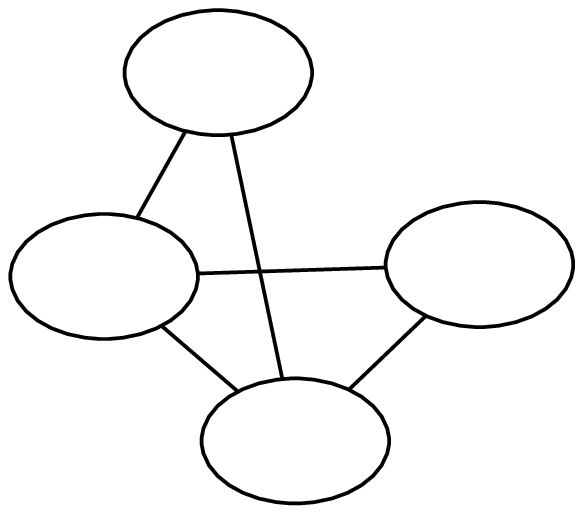}
\end{center}
\subcaption{Support: $999$}
\label{fig: frequet-a}
\end{minipage}
\begin{minipage}{0.19\hsize}
\begin{center}
\includegraphics[width=\hsize]{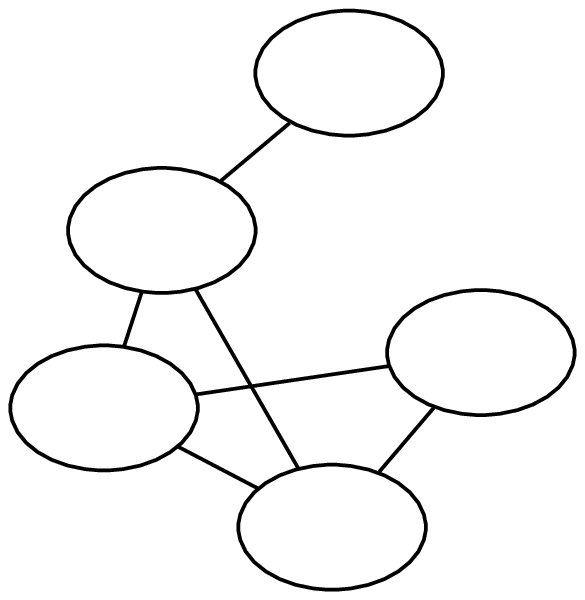}
\end{center}
\vspace{-0.17in}
\subcaption{Support: $999$}
\label{fig: frequet-b}
\end{minipage}
\begin{minipage}{0.19\hsize}
\begin{center}
\includegraphics[width=\hsize]{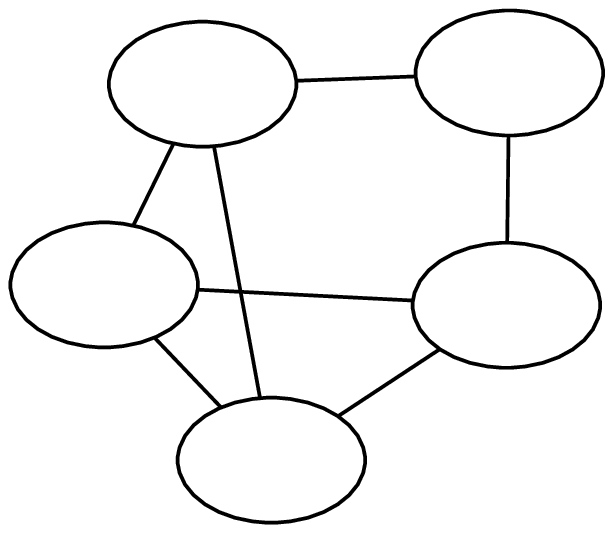}
\end{center}
\subcaption{Support: $962$}
\label{fig: frequet-c}
\end{minipage}
\begin{minipage}{0.19\hsize}
\begin{center}
\includegraphics[width=\hsize]{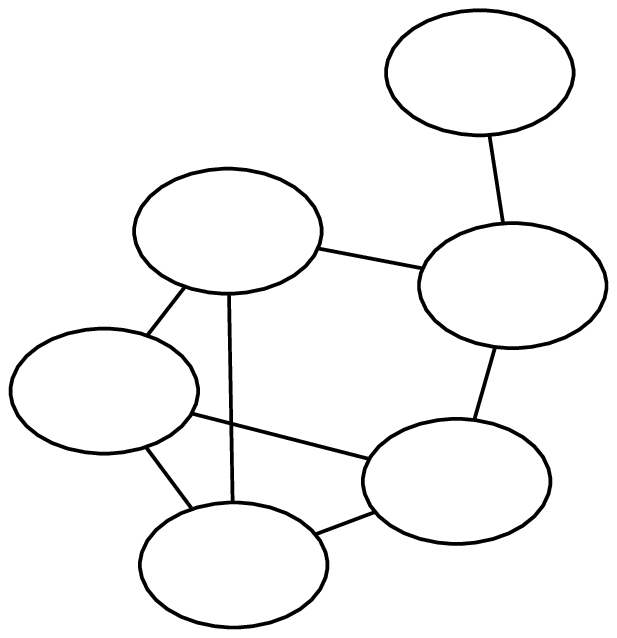}
\end{center}
\vspace{-0.17in}
\subcaption{Support: $959$}
\label{fig: frequet-d}
\end{minipage}
\begin{minipage}{0.19\hsize}
\begin{center}
\includegraphics[width=\hsize]{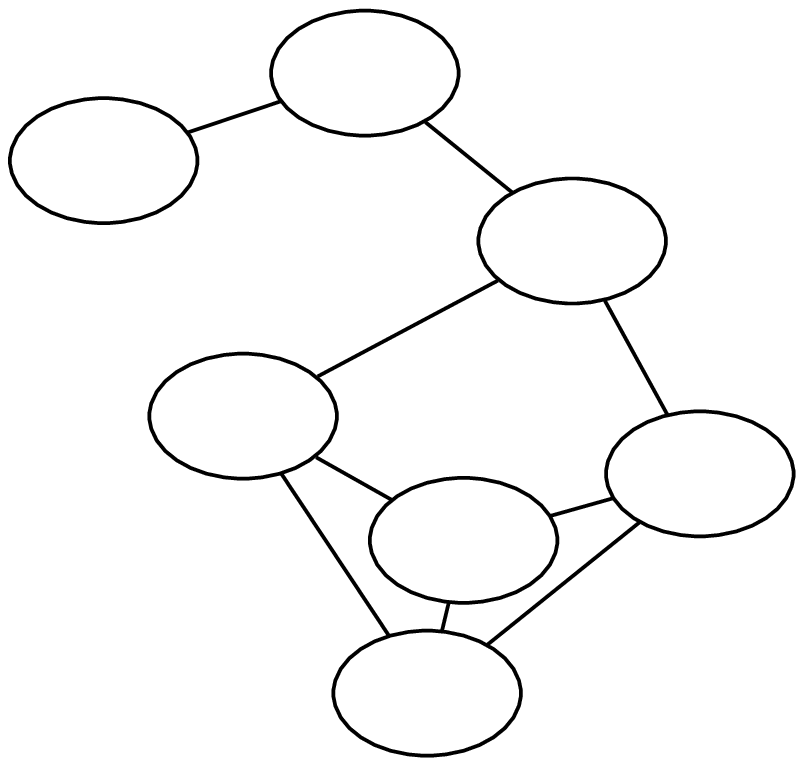}
\end{center}
\vspace{-0.1in}
\subcaption{Support: $959$}
\label{fig: frequet-e}
\end{minipage}
\caption{Frequent subgraphs of the hard distribution for DSATUR.}
\label{fig: frequent}
\end{figure}

We demonstrate how to extract meaningful patterns from the hard graph distribution. Toward this end, we use frequent subgraph mining \citep{AGM, FSG}. This discovers frequent patterns appeared in a database of graphs. We sampled $1000$ graph from the distribution that \textsc{HiSampler} learns for DSATUR. We utilize gSpan \citep{gSpan} to extract frequent subgraphs of them. Figure \ref{fig: frequent} lists the frequent subgraphs that has at least $5$ edges and appears in more than $950$ sampled graphs. Figure \ref{fig: frequet-a} corresponds to the 4-clique highlighted in orange in Figure \ref{fig: brelaz_example}, and Figure \ref{fig: frequet-e} is a subgraph where a small unsolvable graph is connected to a path, which is effective structure as we analyzed in Section \ref{sec: extraction}. It indicates that frequent subgraph mining tools are useful for extracting meaningful structure from the hard graph distribution.

\section{Conclusion} \label{sec:conclusion}

This work tackled the problem of learning the distribution of hard instances using machine learning for the first time. We proposed \textsc{HiSampler} to model the hard instance distribution of graph algorithms. \textsc{HiSampler} is applicable to any graph algorithm without any prior knowledge. We demonstrated the effectiveness of \textsc{HiSampler} using seven algorithms for four graph problems. Furthermore, we showed that hard instances provided insight to analyze and accelerate the algorithm. We also showed that frequent subgraph mining extracts meaningful patterns from the hard graph distribution.

We discuss some future work of this work. Many existing works have tackled molecular generation using deep learning models. Comparing \textsc{HiSampler} with these methods and utilizing them for modeling the hard graph distribution is important future work. Besides, we model the distribution of graphs using adjacency matrices. We do not take isomorphism into account because some algorithms utilize node indices for tie-breaking. However, this limits the effectiveness of \textsc{HiSampler} for algorithms that utilize isomorphism by, for example, restarting with randomization. Modeling symmetry of graphs for such algorithms is promising future work.

\section{Acknowledgments}

This work was supported by JSPS KAKENHI Grant Number 15H01704. MY is supported by the JST PRESTO program JPMJPR165A. We thank Yasuaki Kobayashi and Alessio Conte for discussing about the extensions of our proposed method.

\bibliographystyle{plain}
\bibliography{instance}

\end{document}

%% file: mathdef.tex
\newcommand{\boldA}{{\boldsymbol{A}}}

\newcommand{\boldP}{{\boldsymbol{P}}}

\newcommand{\boldz}{{\boldsymbol{z}}}

\newcommand{\boldtheta}{{\boldsymbol{\theta}}}

